# Technical Report:

# Development of Open Informal Dataset Affecting Autonomous Driving

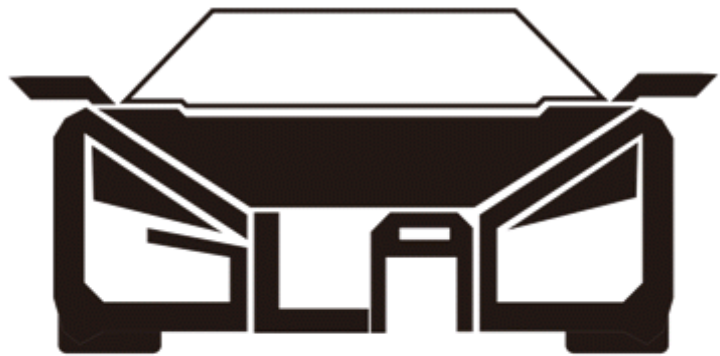

Yong-gu Lee, Sung-jae Lee, Sang-jin Lee, Dong-whan Lee, Tae-seung Baek,

Kyeong-chan Jang, Ho-jin Son, Jin-su Kim

Gwangju Institute of Science and Technology

Gist Laboratory of Autonomous Driving

7th September

# TABLE OF CONTENTS

# 1. SUMMARY

This document is a document that has written procedures and methods for collecting objects and unstructured dynamic data on the road for the development of object recognition technology for self-driving cars, and outlines the methods of collecting data, annotation data, classification criteria of objects, and data processing methods. Before collecting data sets, we designed the data collection method by referring to the KITTI dataset [1]. On-road object and unstructured dynamic data were collected in various environments, such as weather, time and traffic conditions, and additional reception calls for police and traffic controller were collected. Finally, 100,000 images of various objects existing on pedestrians and roads, 200,000 images of police and traffic safety personnel, 5,000 images of police and traffic safety personnel, and data sets consisting of 5,000 image data were collected and built.

# 2. OBJECTS ON-ROAD

## 2.1. DATA COLLECTION

### 2.1.1. METHOD OF DATA COLLECTION

For data collection, two methods were collected, one by attaching a camera to the front of the vehicle and the other by using a camera tripod to capture it in a fixed position. The filming location is 4K (3840x2160) with a resolution of 30FPS. The total length of the collected driving images is approximately 1,600 hours.

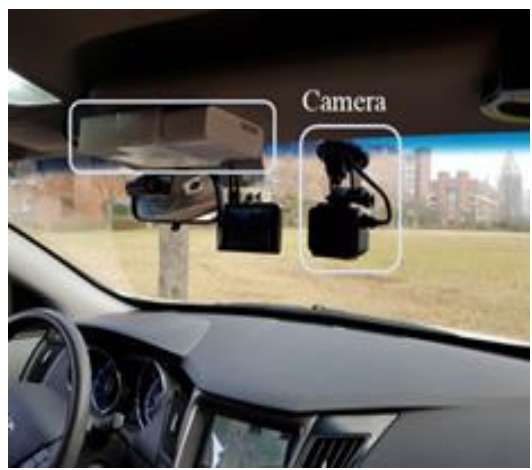

Figure 1 Camera settings for data collection

### 2.1.2. LOCATION OF DATA COLLECTION

As shown in Figure 2, the data were collected in the Gwanghwamun area in Seoul, Seocho-gu in Seoul,

Sejong City, Daegu Metropolitan City, and Gwangju Metropolitan City, and Gwangju Metropolitan City was divided into general roads and campuses within the Gwangju Institute of Science and Technology.

### 2.1.3. ENVIRONMENTS OF DATA COLLECTION

For data diversity, the four seasons of spring, summer, fall, and winter were included, and the shooting weather and time were collected separately.

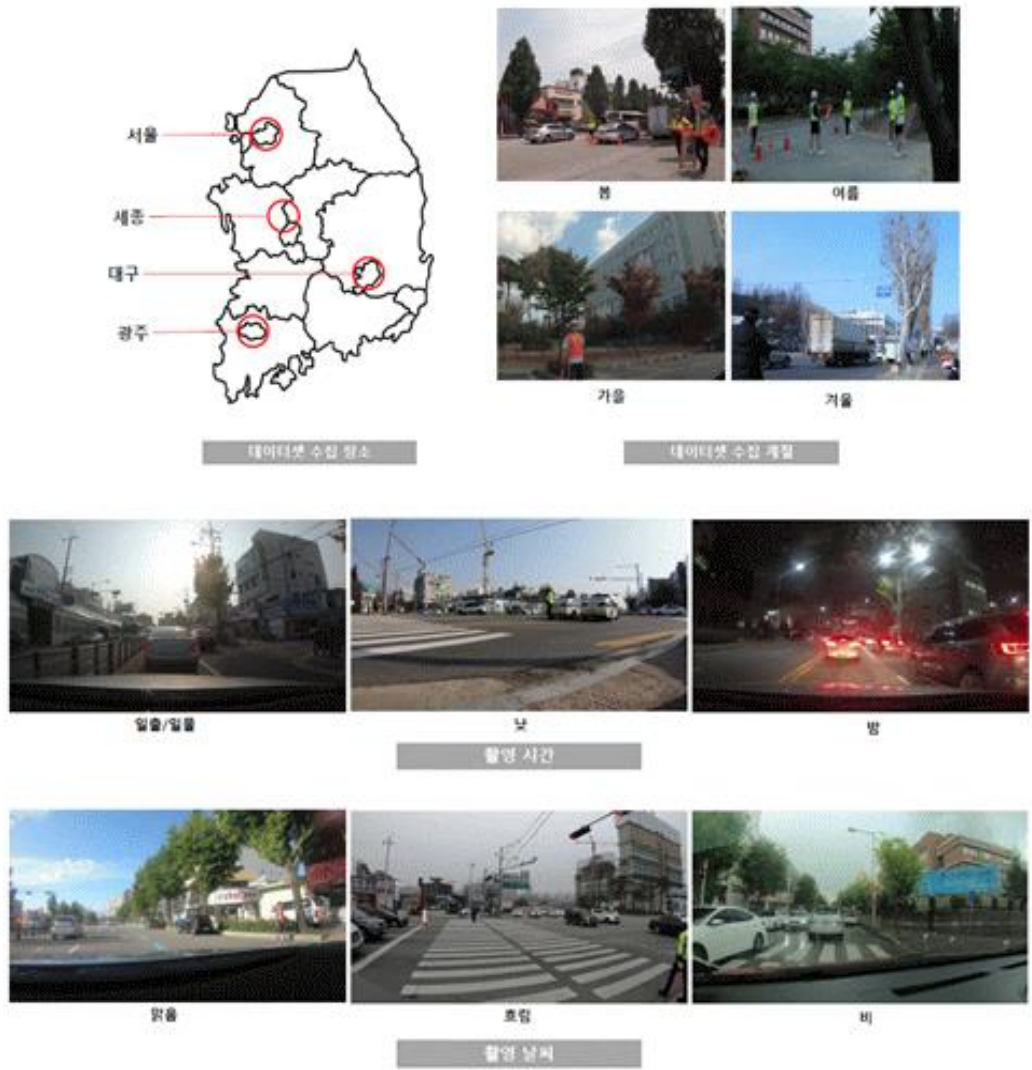

**Figure 2 Data Collection Places and Environments**

## 2.2. ANNOTATION DATA

### 2.2.1. ANNOTATION DATA REPRESENTATION METHOD

Annotation representation of objects present in the image consists of the name of the image file in which the object exists for each object, the classification number according to the classification system of the object, the coordinates of the object representing the image location of the object, and the size of the object, as shown in Figure 3. The classification number of objects consists of Category, Division, and Sub-item, Section, based on factors that affect the driving of vehicles as provided in the Road Traffic Act, the Road Act, and the Ministry of Public Administration and Security Ordinance [2]. (Figure 3) shows Category (pedestrians, cars, road signs, safety signs, traffic lights, etc.) in the manner of annotation data expression and the classification number of objects.

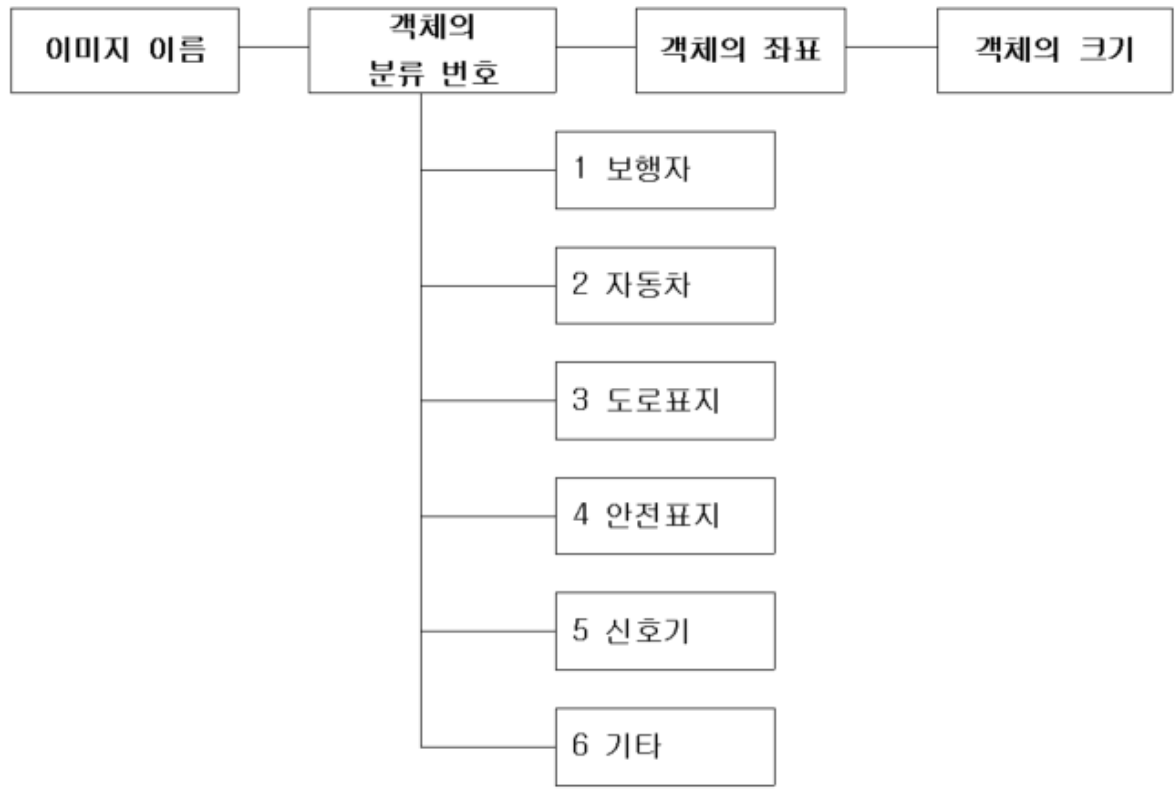

**Figure 3 Categories of annotation data representation and object classification numbers**

## 2.3. OBJECT CLASSIFICATION SYSTEM OF ON-ROAD OBJECTS

### 2.3.1. OBJECT CLASSIFICATION SYSTEM

The object of data on the road consists of pedestrians, cars, road signs, safety signs, and signals that affect the driving of self-driving cars. List the numbers classified by category, division, and section in order to complete the final identification number. The final identification number indicates the type of object for the road data. In addition, in addition to the object classification system presented below, the classification system could be expanded and modified to suit the purpose of the user. The number of detailed object numbers totals 116 types, with a breakdown table. It can be found in the TTA standard [3].

## 2.3.2. OBJECT CLASSIFICATION NUMBER

The objects present in this data set include 60 of the total 116 classifications and are organized as shown in Table 1.

| Category | 분류 번호 | Division | 분류 번호 | Section | 분류 번호 | Small Section | 분류 번호 | 최종 분류 번호 |
|---|---|---|---|---|---|---|---|---|
| 보행자 | 1 | 보행자 | 01 | - | - | - | - | 101 |
| | | 경찰 | 05 | 경찰 | 01 | - | - | 10501 |
| | | | | 경광봉 소지 | 02 | - | - | 10502 |
| | | | | 경찰조끼 착용 | 03 | - | - | 10503 |
| | | | | 경광봉+경찰조끼 | 04 | - | - | 10504 |
| | | 안전요원 | 06 | 경광봉 소지 | 01 | - | - | 10601 |
| | | | | 녹색 조끼 착용 | 02 | - | - | 10602 |
| | | | | 경광봉+녹색 조끼 | 03 | - | - | 10603 |
| | | | | 적색 조끼 착용 | 04 | - | - | 10604 |
| | | | | 경광봉 + 적색 조끼 | 05 | - | - | 10605 |
| 차 | 2 | 자전거 | 02 | 자전거 | 02 | - | - | 20202 |
| | | 자동차 | 03 | 승합차 | 02 | 버스 | 01 | 2030201 |
| | | | | 이륜차 | 05 | - | - | 20305 |
| 도로표지 | 3 | - | - | - | - | - | - | 3 |

| | | | | | | | |
|---|---|---|---|---|---|---|---|
| 안전표지 | 4 | 주의표지 | 01 | - | | - | - | - | 401 |
| | | 규제표지 | 02 | - | - | - | - | 402 |
| | | 지시표지 | 03 | - | - | - | - | 403 |
| | | 보조표지 | 04 | - | - | - | - | 404 |
| | | 노면표지 | 05 | 진행방향 | 01 | 직진 | 01 | 4050101 |
| | | | | | | 좌회전 | 02 | 4050102 |
| | | | | | | 우회전 | 03 | 4050103 |
| | | | | | | 직진, 좌회전 | 04 | 4050104 |
| | | | | | | 직진, 우회전 | 05 | 4050105 |
| | | | | | | 유턴 | 06 | 4050106 |
| | | | | | | 좌회전 및 유턴 | 07 | 4050107 |
| | | | | | | 진행방향 및 방면 직진, 좌회전 | 08 | 4050108 |
| | | | | | | 진행방향 및 방면 우회전 | 09 | 4050109 |
| | | | | | | 비보호 좌회전 | 10 | 4050110 |
| | | | | 금지 | 02 | 직진 금지 | 01 | 4050201 |
| | | | | | | 좌회전 금지 | 02 | 4050202 |
| | | | | | | 우회전 금지 | 03 | 4050203 |
| | | | | | | 직진 및 자회전 금지 | 04 | 4050204 |
| | | | | | | 직진 및 우회전 금지 | 05 | 4050205 |
| | | | | | | 좌우 회전 금지 | 06 | 4050206 |
| | | | | | | 유턴 금지 | 07 | 4050207 |
| | | | | | | 정차 금지 | 08 | 4050208 |

| | | | | 속도 제한 | 01 | 4050301 |
|---|---|---|---|---|---|---|
| | | | 제한 | 03 | 속도 제한(어린이 보호구역 안) | 02 | 4050302 |
| | | | 서행 | 04 | 서행(한글) | 01 | 4050401 |
| | | | | | 서행(그림) | 02 | 4050402 |
| | | | 일시정지 | 05 | - | - | 40505 |
| | | | 양보 | 06 | - | - | 40506 |
| | | | 횡단보도 | 07 | - | - | 40507 |
| | | | | | 횡단보도 예고 | 01 | 4050701 |
| | | | 자전거 | 08 | 자전거 전용도로 | 01 | 4050801 |
| | | | | | 자전거 우선도로 | 02 | 4050802 |
| | | | 보호구역 | 09 | 어린이 보호구역 | 01 | 4050901 |
| | | | | | 노인보호구역 | 02 | 4050902 |
| | | | | | 장애인 보호구역 | 03 | 4050903 |
| | | | 안전지대 | 10 | - | - | 40510 |
| | | | 차로변경 | 11 | - | - | 40511 |
| | | | 오르막경사면 | 12 | - | - | 40512 |
| 신호기 | 5 | 차량 신호등 | 01 | 녹색 | 08 | - | - | 50108 |
| | | | | 적색 | 09 | - | - | 50109 |
| | | 보행자 | 02 | 녹색 등화 | 01 | - | - | 50201 |

|  |  | 신호등 |  | 적색 등화 | 02 | - | - | 50202 |
|---|---|---|---|---|---|---|---|---|
| 정류장 | 6 | 버스 정류장 | 01 | - | - | - | - | 601 |

**Table 1 On-road object classification table**

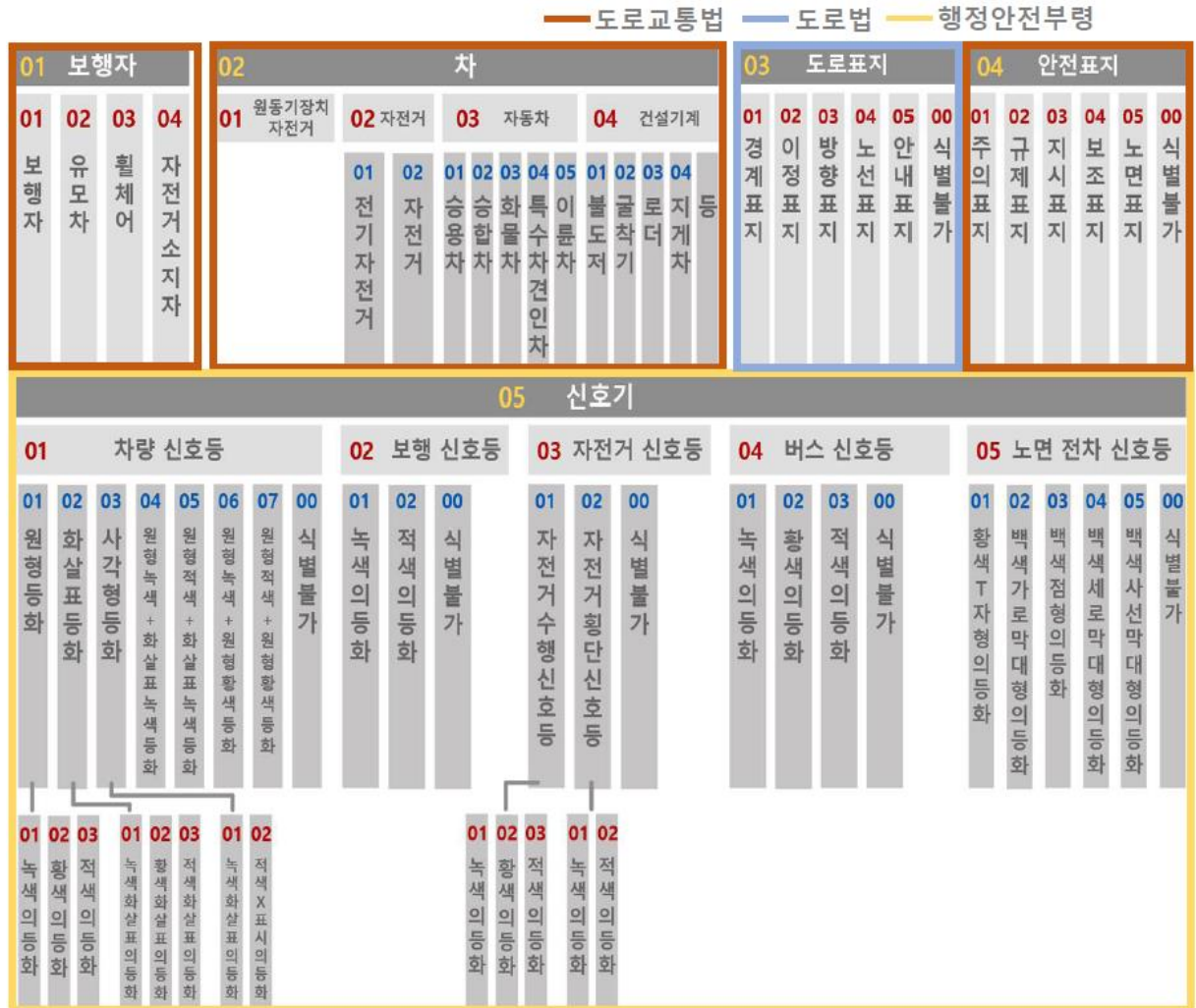

**Figure 4 Classification system of objects on the road**

## 2.4. DATA PROCESSING

### 2.4.1. RAW DATA PROCESSING

In order to process the collected image data into learning data for developing object recognition algorithms, key frames with available objects for learning among the image data must be extracted as images. Through this operation, 300,000 images were extracted from the total 1,600 hours of driving images.

### 2.4.2. DATA INFORMATION PRESENTATION METHOD

Annotation files for image data are required for object recognition algorithms. Annotation files record information (object type, location, and size) to process image files. Each annotation file with annotation information that contains the classification of objects in each image is processed accordingly to that image. Object image · annotation data consists of a filename of the same name, the filename of the image and the labeling format within the text are named as shown in (Figure 5) and have the following meanings.

- Meaning of image filename

: The filename allows you to obtain metadata about the data's filming date and the filming environment.

<Filming date>_<modification status, filming time, weather>_<order of original images>_<frame count per second>_<image order>

- Labeling Format and Meaning

: The labeling file format of the image data collected is '.txt', and has the format shown in Figure 3. Labels in the data are divided into detection labels and tracking labels, which are similar in type to detection labels, but each object is given a unique ID for tracking.

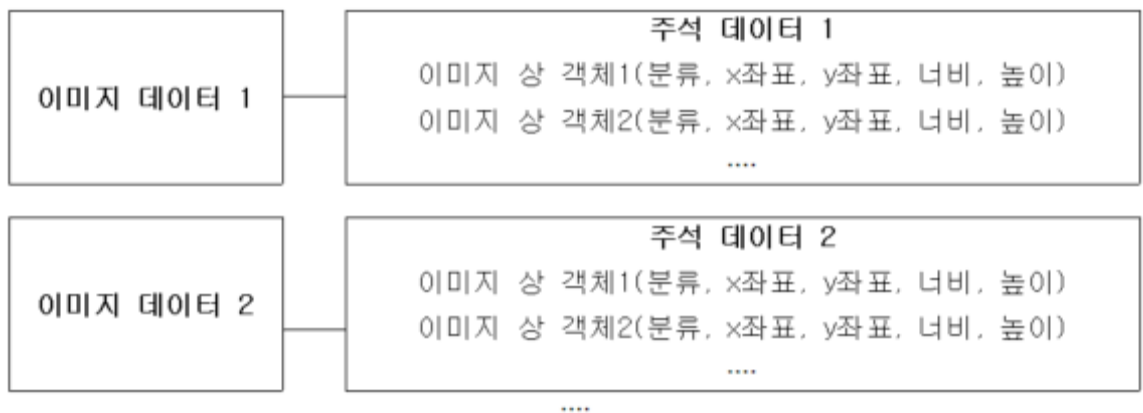

Figure 5 Image data and annotation data representing objects on the image

## 2.4.3 ANNOTATION DATA PRESENTATION METHOD

Identify the existence of objects suitable for the purpose of the user in relation to the image of objects on the road. Pedestrian, signs, bus stops, etc. are in the image of the following example, and classification numbers for objects may be indicated according to the criteria in Table 1. The following (Figure 6) shows an image of annotation information and examples of annotation file.

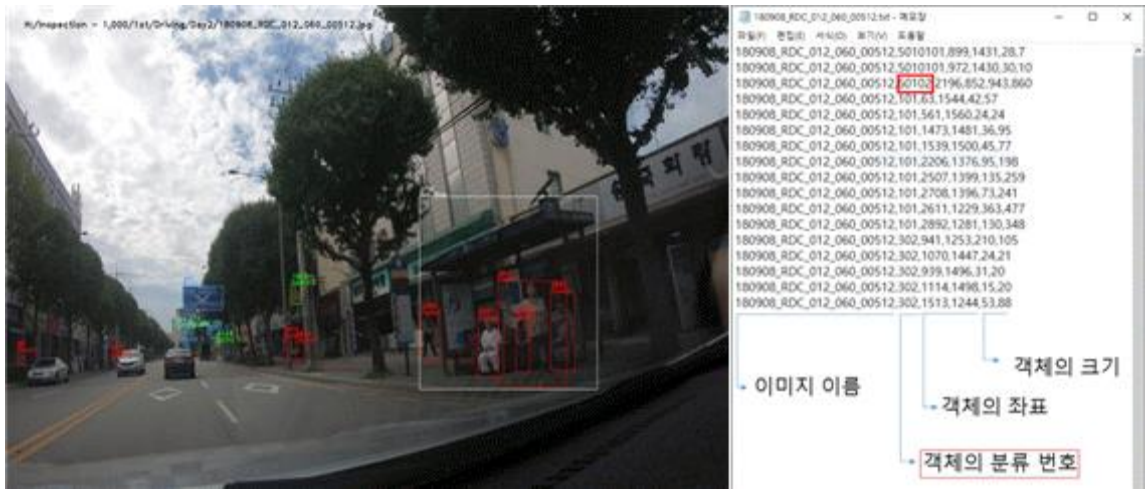

Figure 6 Annotated images (left); Annotated files of images (right)

## 2.4.4. NON-IDENTIFICATION

Image data includes information that is sensitive to privacy, such as a person's facial skin. Therefore, in order to disclose data, un-identification for the protection of personal information should be carried out in accordance with the Personal Information Protection Act [4], so that the face of a person could not be identified by mosaic.

## 2.4.5. INSPECTION AND FINAL COMPLETION

To improve the quality of the data, the processed data set inspected the object misclassification, the object annotation, and the image smudging, and corrected the deficiencies and finally completed the data.

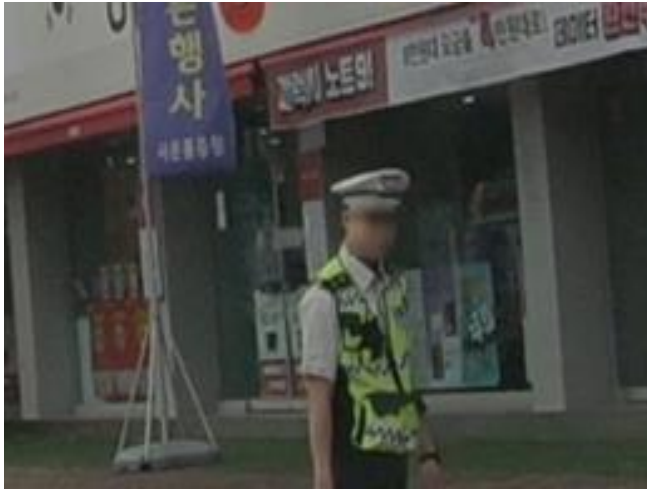

**Figure 7 Non-identification for Personal Information Protection**

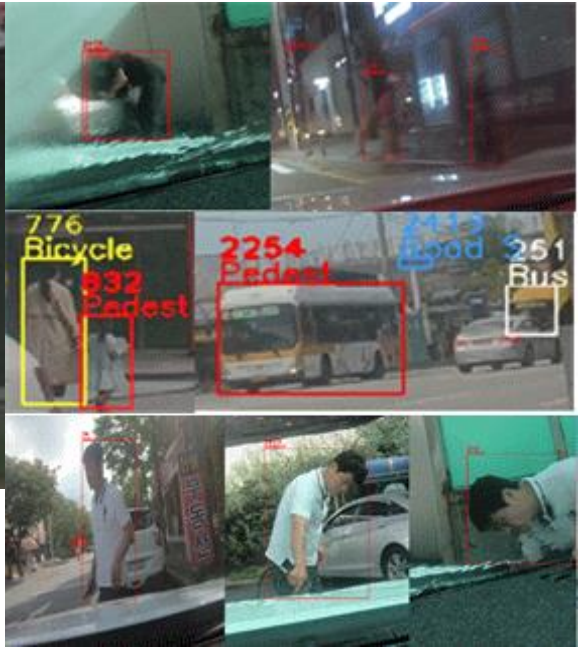

**Figure 8 Data Error Example**

# 3. TRAFFIC CONTROL SIGNAL DATA FROM POLICE OFFICER AND TRAFFIC CONTROLLER

## 3.1. DATA COLLECTION

### 3.1.1. METHOD OF DATA COLLECTION

Traffic control signals from police and traffic controller were largely divided into two types: light sticks and hand signals. For signals using light bars, two methods were collected for data collection: attaching a camera to the front of the vehicle and collecting it while driving the vehicle and taking it from a fixed position using a camera tripod. For reception calls, the method taken using a fixed camera was used. The resolution of the filmed images was 4K (3840x2160) and were taken at 30FPS.

### 3.1.2. LOCATION OF DATA COLLECTION

Filming for data collection was conducted on roads throughout Gwangju, inside the Gwangju Institute of Science and Technology, and inside the Gwangju Children's Transportation Park.

### 3.1.3. ENVIRONMENTS OF DATA COLLECTION

As with the object data on the road, all four seasons were included in the collection for the diversity of the data, and the shooting weather and time were collected equally.

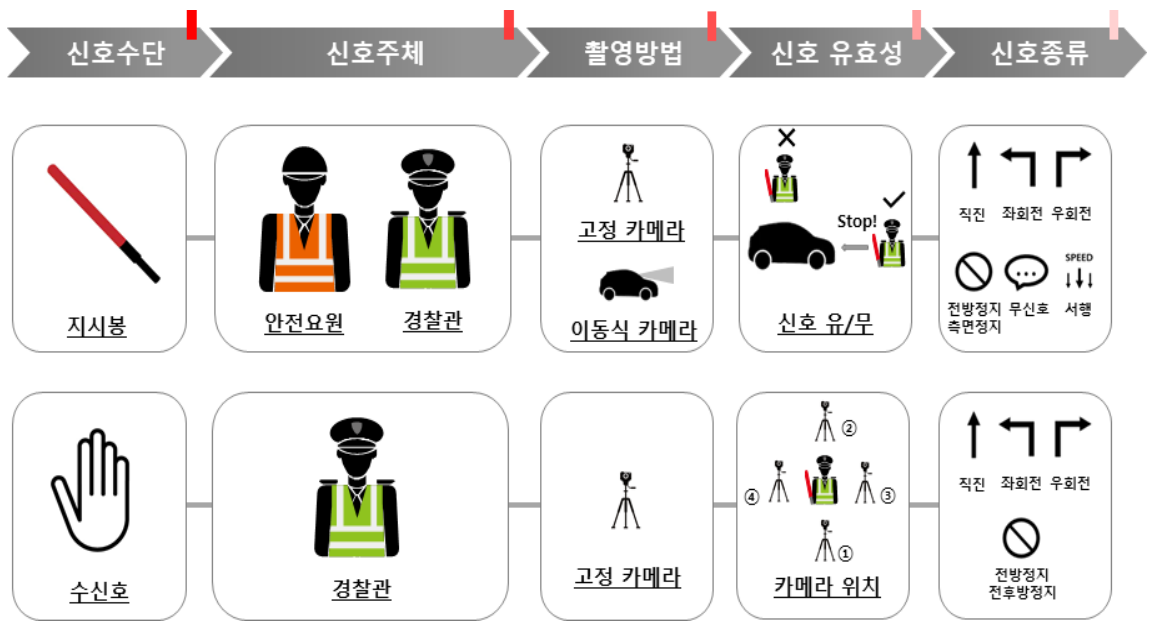

**Figure 9 Traffic Control Signal Data Classification**

## 3.2. ANNOATION DATA

### 3.2.1. ANNOTATION DATA REPRESENTATION METHOD

Annotation information representation of objects present in the image consists of the image file name for each object, the classification number according to the on-road object data classification system, the coordinates of the object representing the image location of the object, and the size of the object, as shown in Figure 10. In addition, in the case of indicator rods, safety indication rods were further classified as safety indication rods No. 701 in addition to classification numbers according to the on-road object data classification system [3]. In the case of a receiving call, the object's attitude information was added, including the information mentioned above.

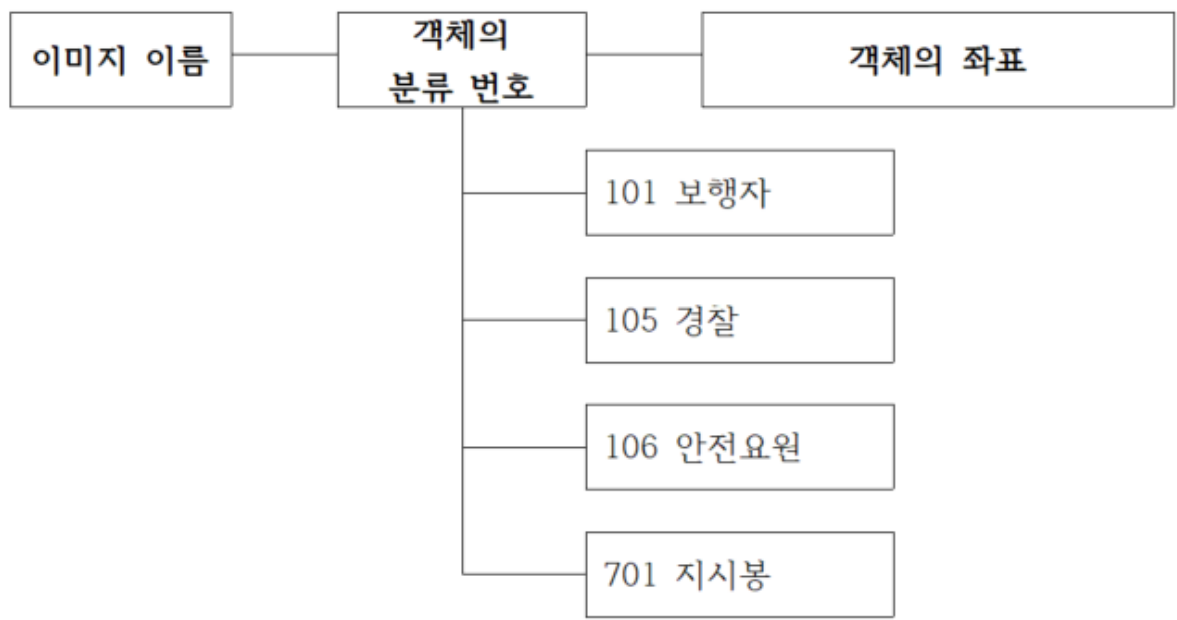

Figure 10 Category of annotation data representation and object classification number of traffic wand

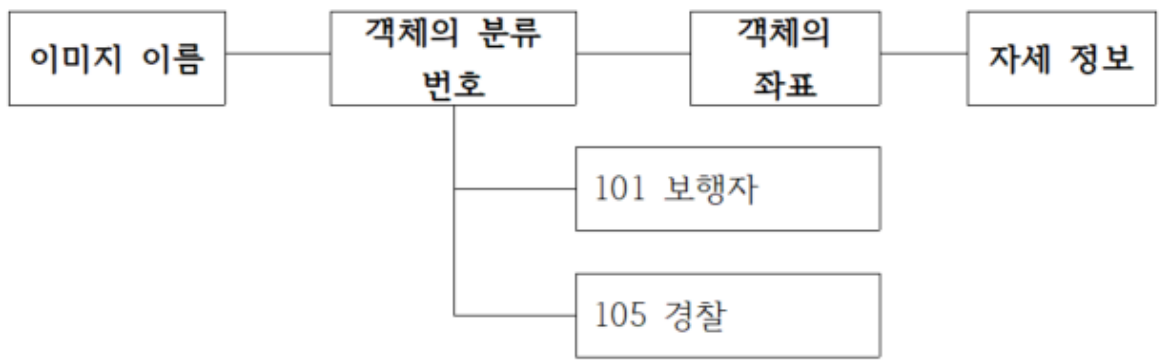

Figure 11 Category of annotation data representation and object classification number in hand signal

## 3.3. TRAFFIC CONTROL SIGNAL CLASSIFICATION SYSTEM

### 3.3.1. USING TRAFFIC WAND

Traffic control signals from police and traffic controller were largely divided into two types: traffic wands signals and hand signals. The purpose of this data is to enable the recognition of signals from signers on a vehicle basis. Signals directed at South Korean vehicles are defined by Article 5 of the Road Traffic Act [2]. According to Article 5 of Korea's Road Traffic Act, light-light-bar signals can be checked by traffic safety personnel. There are a total of three signals specified in the law, but they were further subdivided into a total of seven (Go, Turn left, Turn right, Stop front, Stop side, and Slow) and police and safety personnel were separated from each other to increase awareness at the detection stage. It was also divided into meaningful signals when the receiver signals the camera, or meaningless signals if not.

| 경광봉 신호 종류 | | | | | | | |
|---|---|---|---|---|---|---|---|
| 신호 구분 | 진행 | 좌회전 | 우회전 | 정지 (전방) | 정지 (측방) | 무신호 | 서행 |
| 신호 번호 | 1 | 2 | 3 | 4 | 5 | 6 | 7 |
| 신호의 유효성 | | | | | | | |
| 신호 구분 | 유의미 | | | | 무의미 | | |
| 신호 번호 | 1 | | | | 0 | | |
| 최종 신호 분류 | | | | | | | | | | | | | |
| 신호 구분 | 유의미 | | | | | | | 무의미 | | | | | |
| 신호 번호 | 1 | | | | | | | 0 | | | | | |
| 최종 분류 번호 | 진행 | 좌회전 | 우회전 | 정지 (전방) | 정지 (측방) | 무신호 | 서행 | 진행 | 좌회전 | 우회전 | 정지 (전방) | 정지 (측방) | 무신호 | 서행 |
| | 11 | 12 | 13 | 14 | 15 | 16 | 17 | 01 | 02 | 03 | 04 | 05 | 06 | 07 |

Table 2 Classification criteria and classification number of traffic wands signs

| 신호 구분 | 동작 예시 |
|---|---|
| 진행 | |
| 좌회전 | |
| 우회전 | |
| 정지(전방) | |
| 정지(측방) | |
| 무신호 | |
| 서행 | |

**Table 3 Examples of traffic wands signals**

## 3.3.2. HAND SIGNAL

As with signals using traffic wand, reception was also aimed at recognizing signals from receivers. According to Article 9 of the Enforcement Rules of the Road Traffic Act [2], the types of signals and the method of display are prescribed, and according to the police officers' signal culture plan provided by the Central Police School , the classification of motion is largely divided into three types: progress signal, rotation signal and stop signal, and there are a total of 16 subdivided actions. However, considering the recognition of signals at the point of time of the vehicle, data were collected by determining five actions that could be recognised at the point of time of the vehicle, except that 16 signals had overlapping behaviors depending on the location and timing of the vehicle. The five movements consist of a forward-to-rearward progression, a forward-to-left progression, a forward-to-right progression, a forward-to-back stop and a forward-to-back simultaneous stop. Each signal represented a direction based on the signal (police). For example, in the case of a forward-to-rearward progression, it means that the forward vehicle of the signer is going straight after receiving the signal.

| 수신호 종류 | | | | | |
|---|---|---|---|---|---|
| 신호 구분 | 전방에서 후방으로 | 전방에서 좌측으로 | 전방에서 우측으로 | 전방 정지 | 전·후방 동시 정지 |
| 신호 번호 | 1 | 2 | 3 | 4 | 5 |

Table 4 Classification criteria and classification number for hand signals

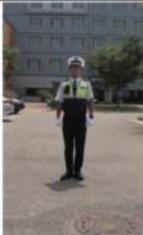

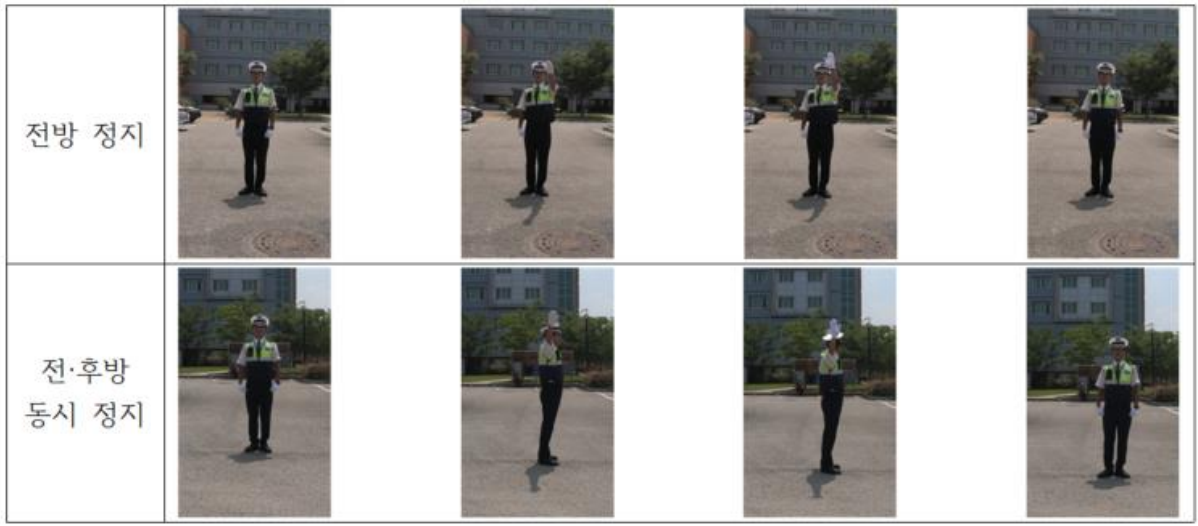

**Table 5 Examples of actions of hand signals**

## 3.4. DATA PROCESSING

### 3.4.1. RAW DATA PROCESSING

  The collected image data has been converted into images (.jpg), and one image contains one action. The image is converted to 30 images per second and consists of an average of 30 images (1 second) per movement for traffic wand signals, and an average of 90 images (3 seconds) per motion for hand signals. All actions were divided from the point at which the action began to the moment when the action was completed and returned to its place.

### 3.4.2. DATA INFORMATION PRESENTATION METHOD

  Traffic control signal recognition algorithm requires annotation files on image data. Annotation files record information (object type, location, and size) to process image files. Each annotation file with annotation information containing the classification of objects in each image is processed accordingly to that image. Object image · annotation data consists of a filename of the same name, the filename of the image and the labeling format within the text are named as shown in Figure 3, and have the following meanings.

- Image file name naming standard

| 구분 | | | 정보 | 표시 정보 |
|---|---|---|---|---|
| 일자 | | | 촬영일자 | 년,월,일 |
| 환경 | 시간 | | 낮 | D |
| | | | 일몰, 일출 | S |
| | | | 밤 | N |
| | 날씨 | | 맑음 | F |
| | | | 흐림 | C |
| | | | 비 | R |
| 동영상 번호 | | | - | 4자리 수 |
| 복장 정보 | | | 초록 조끼 | FG/OG |
| | | | 붉은 조끼 | FR/OR |
| 촬영 방법 | | | 동적 촬영 (차량 주행) | SG |
| | | | 정적 촬영 (삼각대 고정) | DG |
| 신호 번호 | 지시봉 신호 번호 | 유의미 | | 01~07 |
| | | 무의미 | | 11~17 |
| 동영상 동작 세트 번호 | | | 동작 세트 번호 | 6자리 수 |
| 이미지 순서 | | | 신호별 이미지 번호 | 2자리 수 |
| 예시 | | | 20200717_DC_0011_OG_SG_11_000007_01.jpg | |

**Table 6 Image file name naming standard (Traffic wand signals)**

| 구분 | | 정보 | 표시 정보 |
|---|---|---|---|
| 일자 | | 촬영일자 | 년,월,일 |
| 환경 | 시간 | 낮 | D |
| | 날씨 | 맑음 | F |
| | | 흐림 | C |
| 동영상 번호 | | 촬영 동영상 번호 | 4자리 수 |
| 복장 정보 | | 초록 조끼 | OG |
| 촬영 방법 | | 정적 촬영 (삼각대 고정) | SG |
| 신호 번호 | | 수신호 동작 번호 | 01~15 |
| 카메라 위치 | | 카메라 위치정보 | 1~4 |
| 이미지 순서 | | 신호별 이미지 번호 | 3자리 수 |
| 예시 | | 20200717_DS_0001_OG_SG_01_1_001.jpg | |

**Table 7 Image file name naming standard (Hand signals)**

- Labeling format and meaning of labeling

: The labelling file format of the collected image data is '.txt', and in the case of the traffic wand signals, the annotation data contained information about the objects (walkers, police, and safety personnel) and the indicator bars that exist on the image. For the sign data, the behavior number of the posture, including the object's classification number and location information, is included additionally.

- Hand signals position action number

: Hand signals position action number (01–05) and the number of the delimited operation (00–03) are expressed in combination with the two numbers. For example, for signals from front to rear, the action number is 01 and, if the identification action number indicates the attitude information of the image data, it can be expressed as 0100, 0101, 0102, 0103, 0100. All posture information starts with the starting position, 00, and ends with the finishing position, 00. For the signal data, the annotation information was entered by separating it from the middle point between actions, since it consists of approximately 90 images. For example, if the number of images going from 1 to 2 is 20, the first 10 are represented as 00 and the second 10 as 01.

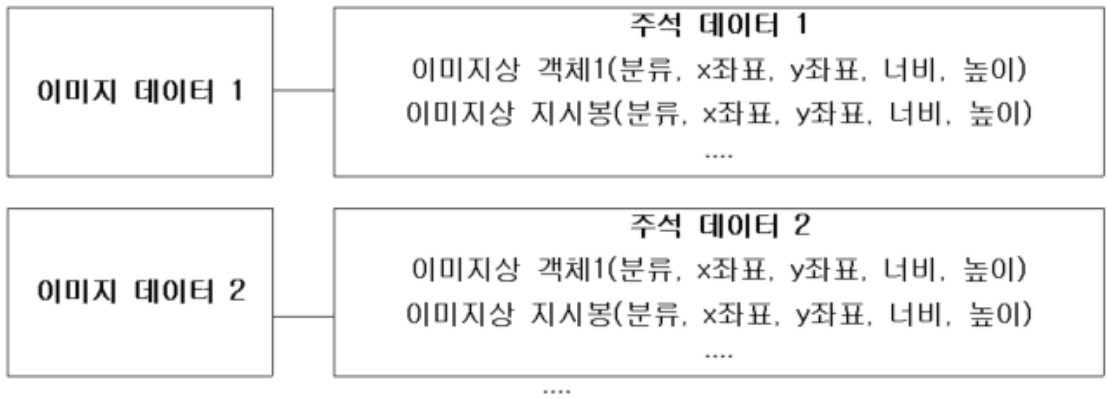

**Figure 12 Image data using traffic wands and annotation data representing objects**

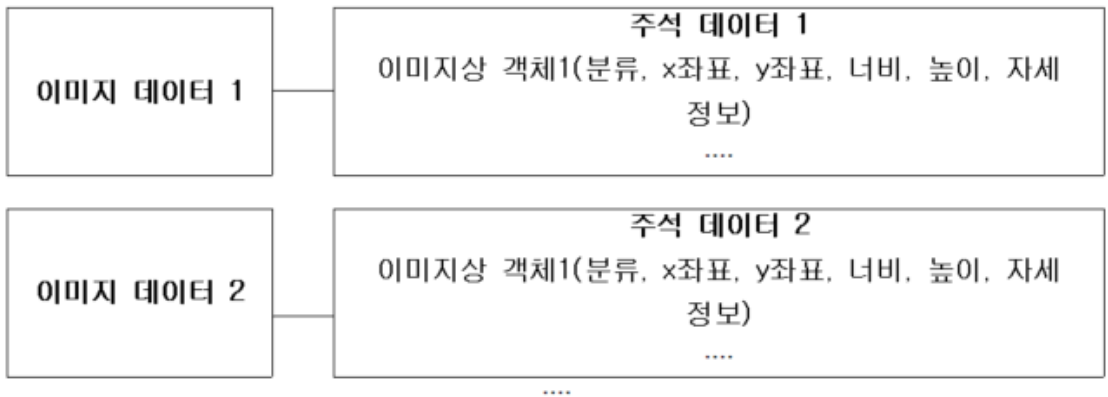

**Figure 13 Image data using hand signals and annotation data representing objects**

| 신호 구분 | 동작 예시 | | | | |
|---|---|---|---|---|---|
| | 1동작 | 2동작 | 3동작 | 4동작 | 5동작 |
| 전방에서 후방으로 | | | | | |
| 동작 번호(01) | 0100 | 0101 | 0102 | 0103 | 0100 |
| 전방에서 좌측으로 | | | | | - |

| 동작 번호(02) | 0200 | 0201 | 0202 | 0200 | - |
|---|---|---|---|---|---|
| 전방에서 우측으로 | 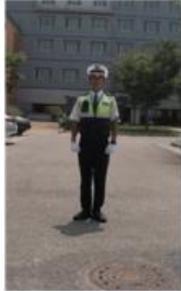 | 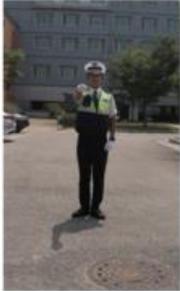 | 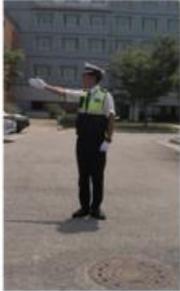 | 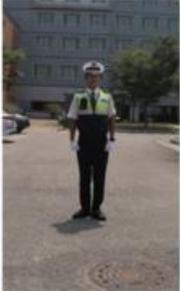 | - |
| 동작 번호(03) | 0300 | 0301 | 0302 | 0303 | - |
| 전방 정지 | 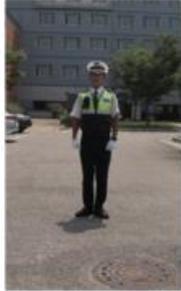 | 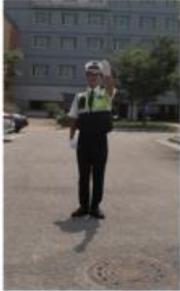 | 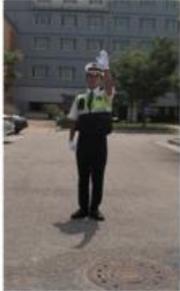 | 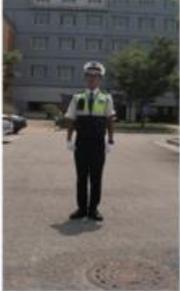 | - |
| 동작 번호(04) | 0400 | 0401 | 0402 | 0403 | - |
| 전·후방 동시 정지 | 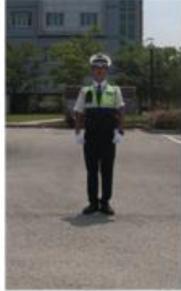 | 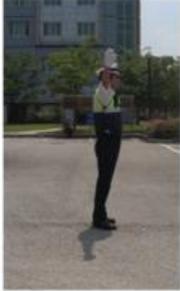 | 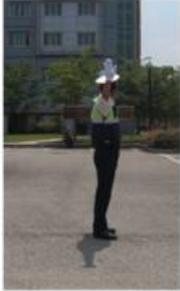 | 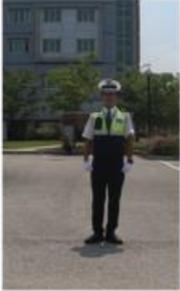 | - |
| 동작 번호(05) | 0500 | 0501 | 0502 | 0503 | - |

Figure 14 Hand signals action number classification

### 3.4.3. ANNOTATION DATA PRESENTATION METHOD

Classification numbers for objects on the image may be indicated according to the criteria in [Table. The following (Figure 14) shows an image of annotation information and an example of an annotation file.

### 3.4.4. DATA HIERARCHY

The data structure is largely divided into signal data (police, traffic controller) using the indicator bar and signal data from the police, and sub-items include a folder with image files and a text file item with annotation information for that image. Depending on the filming method, fixed filming was divided into static, while filming while driving was divided into a total of seven signals, and Valid and Invalid were classified according to the validity of the signal. Finally, one action was organized so that several images and annotation files were contained in a single folder. In the case of reception calls, a total of five signals were divided, and one action was similarly organized to contain several images and annotation files in a single folder.

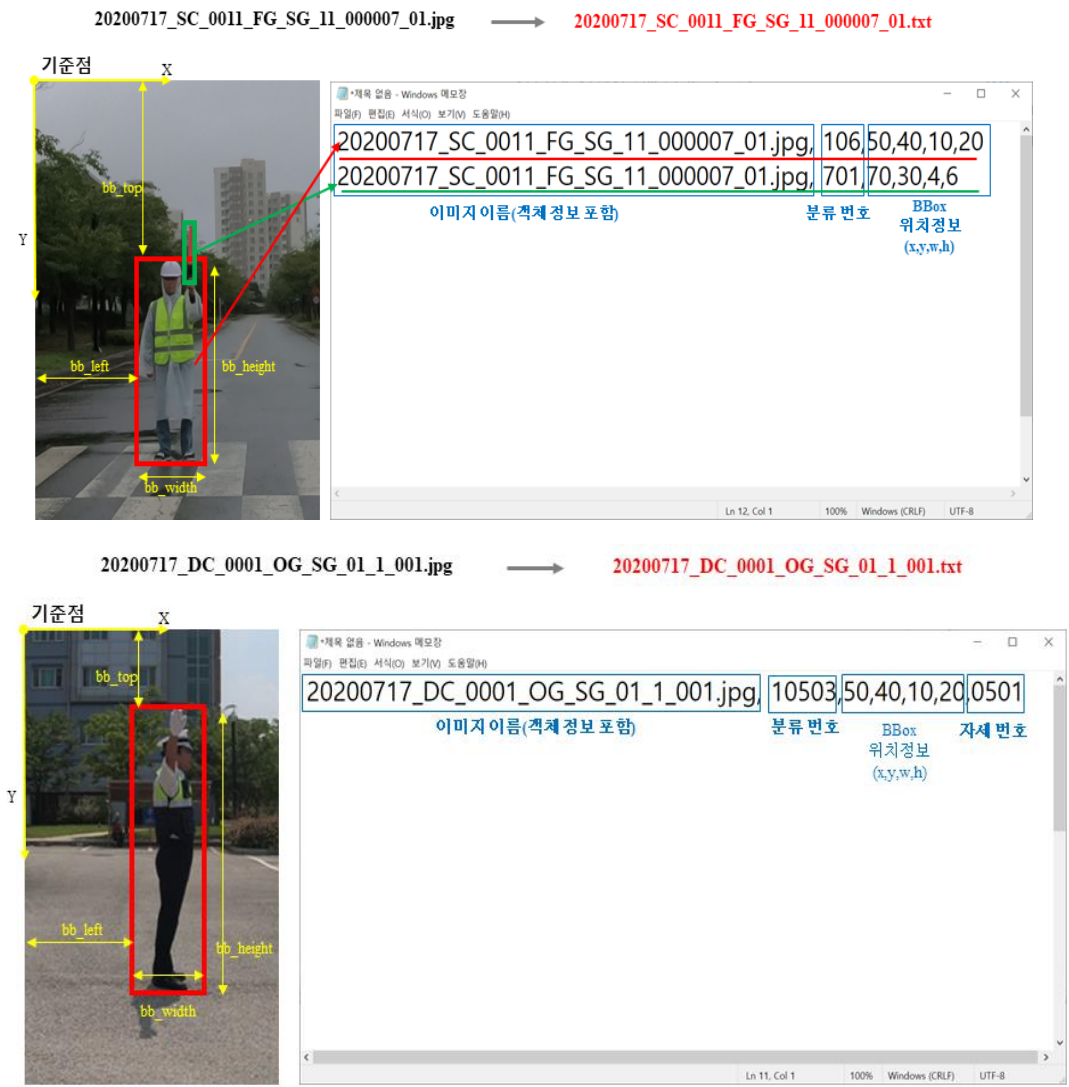

Figure 15 Example of how annotation data should be written

```
• Signal Recognition Datasets/:
    • {Foremen, Police_wand}/
        • {Img, label}/
            • Edited/
                • {static, dynamic}/
                    • {Invalid, Valid}/
                        • {Go, Turn_left, Turn_right, Stop_front, Stop_side, No_signal, Slow}/
                            • Sequence folder/
                                • {Img(.jpg), label(.txt)}

• Police_hand/
    • {Img, label}/
        • Edited/
            • Static/
                • {01,02,03,04,05}/
                    • Sequence folder/
                        • {Img(.jpg), label(.txt)}
```

Figure 116 Data hierarchy

### 3.4.5. NON-IDENTIFICATION

As with the object data on the road, unidentifiable information protection was conducted.

### 3.4.6. INSPECTION AND FINAL COMPLETION

Similar to on-road object data, inspections were conducted to improve data quality.

# 4. REFERENCES

[1] A. Geiger, P. Lenz and R. Urtasun, "Are we ready for autonomous driving? The KITTI vision benchmark suite," 2012 IEEE Conference on Computer Vision and Pattern Recognition, Providence, RI, 2012, pp. 3354-3361, doi: 10.1109/CVPR.2012.6248074.

[2] The Road Traffic Act, vol. 5, no. 9. 2020. Road Act. 2020.

[3] Telecommunications Technology Association, "Object Classification System of On-road Data required for Object Recognition Technology of Autonomous Vehicle", 2019.

[4] Enforcement Decree of the Personal Information Protection Act, vol. 4, no. 15. 2020.